\documentclass{article} % For LaTeX2e
\usepackage[preprint]{colm2026_conference}
\usepackage{booktabs}
\usepackage{pifont}
\usepackage{microtype}
\usepackage{hyperref}
\usepackage{url}
\usepackage{booktabs}
\usepackage{graphicx}
\usepackage{booktabs}
\usepackage{amsmath}
\usepackage{dblfloatfix}
\usepackage{wrapfig}
\usepackage{enumitem}
\usepackage[most]{tcolorbox}
\usepackage{multirow}
\definecolor{lightgrayrow}{RGB}{225,225,225}
\definecolor{lightpurplerow}{RGB}{228,212,245}
\usepackage{subcaption}
\usepackage{booktabs}
\usepackage{pifont}
\newcommand{\cmark}{\ding{51}}
\newcommand{\xmark}{\ding{55}}
\usepackage{lineno}
\usepackage[table]{xcolor}
\definecolor{lightgrayrow}{gray}{0.94}
\definecolor{lightpurplerow}{RGB}{245,239,255}
\definecolor{darkblue}{rgb}{0, 0, 0.5}
\definecolor{lightpurplerow}{RGB}{250,246,255}
\hypersetup{colorlinks=true, citecolor=darkblue, linkcolor=darkblue, urlcolor=darkblue}

\title{Reason Before You Retrieve: Agentic Planning for Multimodal RAG}

\author{
Tianyu Yang\textsuperscript{1,2}, 
Shir Simon\textsuperscript{1}, 
Zhenzhen Li\textsuperscript{1}, 
Minhao Cheng\textsuperscript{3}, 
Xiangliang Zhang\textsuperscript{2} \\
\textsuperscript{1}Bosch AI Research Center \\
\textsuperscript{2}University of Notre Dame \\
\textsuperscript{3}Pennsylvania State University \\
\texttt{tyang4@nd.edu, xzhang33@nd.edu}  \\
\texttt{Shir.Simon@us.bosch.com}
}

\begin{document}

\ifcolmsubmission
\linenumbers
\fi

\maketitle

\begin{abstract}
Multimodal retrieval-augmented generation (mRAG) aims to answer image-text queries with external knowledge, but most existing systems still retrieve directly from raw multimodal input over a flat evidence space. This design often struggles with two key challenges: the retrieval target is under-specified because the question intent must be grounded to the correct visual referent, and the search space is weakly structured, forcing semantically distinct evidence to compete in a single global ranking step. We propose MM-R2, a multimodal agentic retrieval framework that reasons before retrieval by explicitly modeling both what to retrieve and where to search. MM-R2 first constructs an intent-grounded retrieval state from the image-question pair, capturing the information need, grounded referent, and retrieval constraints. It then performs retrieval over a structured KnowledgeMap, where the agent selects relevant retrieval units before issuing grounded queries within them. To enable this capability, we build MM-R2-Traj, a large-scale trajectory dataset of multi-step retrieval processes, and adopt a two-stage post-training strategy with supervised fine-tuning and GRPO. Experiments on Infoseek and Encyclopedic VQA datasets show that MM-R2 substantially outperforms strong baselines on answer accuracy while also yielding more interpretable and verifiable retrieval trajectories.
\end{abstract}

\maketitle
\begingroup
\renewcommand{\thefootnote}{\fnsymbol{footnote}}
\footnotetext[2]{Work done during an internship at Bosch AI Research Center.}
\endgroup

\section{Introduction}
\label{sec:intro}
% \vspace{-3mm}
Recent advances in multimodal large language models~\citep{chen2022murag,hurst2024gpt,team2023gemini,bai2025qwen2,li2024llava,chen2024internvl} have spurred growing interest in answering image--text queries by retrieving from large-scale knowledge bases. Multimodal retrieval-augmented generation (mRAG)~\citep{chen2022murag,hu2023reveal,chen2024mllm} has thus emerged as a promising paradigm for grounding generation with external evidence. While the prevailing ``retrieve-then-generate'' pipeline works well in single-modal RAG---where the query and corpus share the same modality and retrieval intent is implicitly aligned---it often struggles in multimodal settings that require joint reasoning over visual content, textual evidence, and their cross-modal interactions.

The first limitation is \emph{retrieval-intent ambiguity}. In multimodal QA, the text question defines the information need, while the image grounds the particular object, scene, or visual instance to which this need refers. Moreover, the required evidence may span multiple levels of abstraction, ranging from perceptual facts (e.g., objects, attributes, and actions) to relational structures and implicit contextual knowledge (e.g., geographic or historical information). Existing mRAG systems often bypass such retrieval-intent  analysis. Some retrieve independently from textual and visual channels and fuse results afterward~\citep{suri2025visdom,tian-etal-2025-core}, while others collapse multimodal input into a text-space query~\citep{long2024generative}. As a result, retrieval tends to drift toward textual priors or generic caption-like matching, producing evidence that appears relevant in one context but is insufficiently grounded to the actual image--question pair. 

The second limitation stems from retrieval over a flat and weakly structured search space, where semantically diverse evidence is embedded into a single representation space and retrieved via a global similarity function. Many mRAG systems directly select the top-ranked evidence from the pool ~\citep{long2024generative,zhao2025funnelrag,xiao2025graphrag,yang2025omgm},  often favoring generic but loosely related content over more precise, domain-specific evidence. 
%Even when the information need is roughly captured, many mRAG systems still retrieve from a single monolithic evidence pool~\citep{long2024generative,zhao2025funnelrag,xiao2025graphrag,yang2025omgm}. As a result, evidence from semantically distinct  regions of the corpus is forced to compete within one global ranking step, rather than being routed to an appropriate subspace first. This makes multi-hop, multi-aspect, and context-sensitive evidence difficult to retrieve reliably, and often causes relevant evidence to be diluted or crowded out by globally similar but locally mismatched candidates. We refer to this failure mode as \emph{search-space entanglement}, where \emph{where to search} is not modeled as an explicit decision but is instead left as an implicit by-product of similarity scoring.

These limitations highlight the need to reframe  multimodal retrieval as \emph{reason before you retrieve}. This intuition is loosely consistent with  {Schema Theory} in cognitive science~\citep{arbib1992schema}, which posits that humans interpret new inputs by activating structured knowledge frameworks that guide attention, memory access, and inference. Analogously, multimodal retrieval should begin by transforming the image--question pair into a structured, intent-grounded retrieval state, which then activates appropriate regions of the knowledge space and constrains subsequent evidence search.

\begin{figure*}[t]
\centering
\includegraphics[width=1\textwidth]{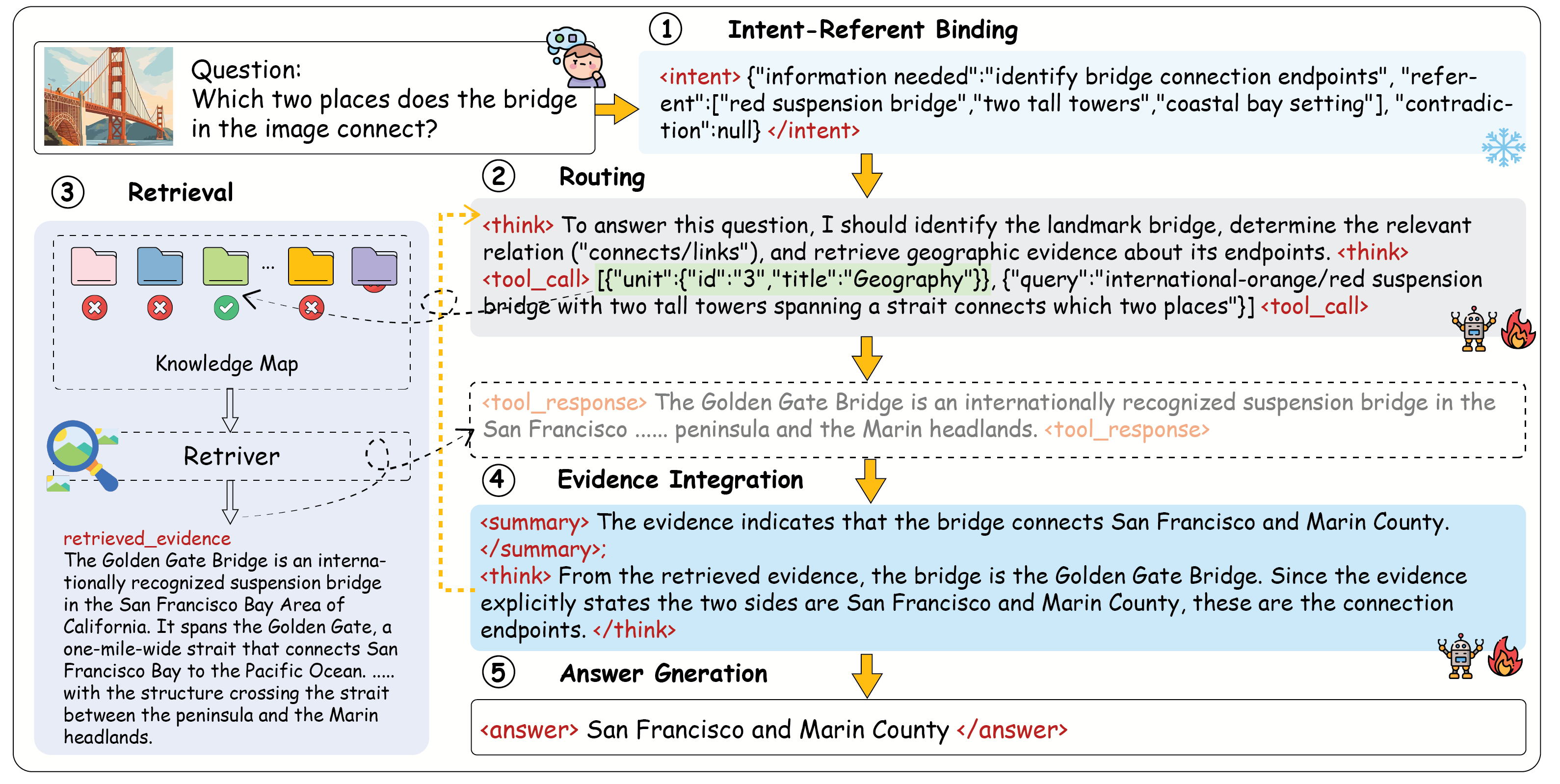}
% \vspace{-9pt}
\caption{\textbf{Overview of our MM-R2 pipeline.} Given an image-question pair, MM-R2 agent first constructs an intent-grounded retrieval state, then reasons over which KnowledgeMap unit(s) to access, retrieves evidence within the selected unit(s), summarizes the retrieved evidence, and determines whether additional retrieval is needed or whether sufficient evidence has been collected to generate the final answer.
} 
% % \vspace{-20pt}
\label{fig1}
\end{figure*}

Building on this idea, we propose \emph{MM-R2} (\textbf{M}ulti\textbf{M}odal \textbf{R}eason-to-\textbf{R}etrieve), a multimodal agentic retrieval framework that  reasons about both \emph{what to retrieve} and \emph{where to search} before retrieval. MM-R2 first employs an \emph{Intent-Referent Binding Module} to infer a \emph{structured retrieval state} from the image--question pair, explicitly modeling the information need, the grounded visual referent, and task-specific constraints. This step resolves retrieval intent prior to search, reducing query drift and providing a structured retrieval state from which the agent can decide where to search.
%an interpretable interface for downstream decisions.
To support this decision, we organize the corpus into semantically coherent \emph{retrieval units} (as a  \emph{KnowledgeMap}), enabling the agent to first select relevant units and then execute grounded queries within them. In contrast to prior mRAG systems that retrieve evidence through a single-step global search, MM-R2 decomposes retrieval into routing and within-unit evidence acquisition. Beyond improving retrieval precision, this decomposition enables a more interpretable retrieval process by allowing inspection of the inferred intent,   selected retrieval units, and   acquired evidence separately.
%MM-R2 further introduces a structured \emph{KnowledgeMap} to make \emph{where to search} an explicit decision. Instead of retrieving from a flat evidence pool, we organize the corpus into semantically coherent retrieval units and allow the agent to first select relevant units, followed by grounded query execution within them. Retrieval is thus decomposed into routing and within-unit evidence acquisition, rather than a single global search. This design yields a more controllable and interpretable retrieval process, particularly for queries requiring cross-region, multi-step, or context-sensitive evidence gathering.

To teach the agent to infer \emph{where to search} from \emph{what to retrieve} (i.e., to map a structured retrieval state to a relevant retrieval unit),  we construct \emph{MM-R2-Traj}, a large-scale trajectory dataset of grounded multi-step retrieval processes, and adopt a two-stage post-training pipeline.
%To learn such behavior, we construct \emph{MM-R2-Traj}, a large-scale trajectory dataset of grounded multi-step retrieval processes, and adopt a two-stage post-training pipeline. 
We first perform supervised fine-tuning (SFT) to train the agent perform structured retrieval behaviors, including unit selection, grounded query generation, evidence summarization, and answer generation. We then further optimize the agent policy with GRPO to strengthen both retrieval-intent reasoning and retrieval-space planning under task-level rewards. Because MM-R2 explicitly exposes intermediate retrieval decisions, we also complement standard answer evaluation with lightweight process-level analysis.

Our contributions are summarized as follows:
\begin{itemize}[leftmargin=*,itemsep=2pt]
    \vspace{-3mm}\item Our proposed MM-R2 is a novel multimodal agentic retrieval framework that explicitly reasons about both the retrieval target and the retrieval space before retrieval, making "what to retrieve" and "where to search" explicit and interpretable.
    \vspace{-1mm}\item To prepare the agent to reason, we design an \emph{Intent-Referent Binding Module} and a structured \emph{KnowledgeMap} to  provide an explicit decision state and a structured search space. 
    \vspace{-3mm}\item The agents trained with MM-R2-Traj are evaluated on multimodal QA benchmarks and achieve SOTA performance.
    Also, we conduct process-level analysis of retrieval behavior, demonstrating that MM-R2 yields more accurate routing, more faithful evidence use, and more transparent retrieval trajectories.

\end{itemize}
\section{Related Works}
% % % \vspace{-3mm}

\noindent
\textbf{Agentic LLM.} Agentic approaches~\citep{schick2023toolformer,nakano2022webgptbrowserassistedquestionansweringhuman,karpas2022mrklsystemsmodularneurosymbolic,patil2023gorillalargelanguagemodel} cast LLMs as planners that interleave reasoning with tool use and produce auditable traces, as in ReAct~\citep{yao2022react} and Toolformer~\citep{schick2023toolformer}. In multimodal settings~\citep{chen2023llavainteractiveallinonedemoimage,zheng2024gpt4visiongeneralistwebagent,press2023measuringnarrowingcompositionalitygap}, systems such as Visual ChatGPT~\citep{wu2023visual}, MM-REACT~\citep{yang2023mm}, and HuggingGPT~\citep{shen2023hugginggpt} show the benefits of planning and tool routing. Our framework does not rely on multi-tool routing; instead, it uses a single RAG tool. It remains agentic through an iterative think--act process, but introduces a pre-coupling step that grounds retrieval intent from multimodal input before planning and action.

% % % \vspace{-1mm}
\noindent
\textbf{Multimodal RAG.} Multimodal RAG (mRAG) extends knowledge-based VQA and multimodal LLMs by integrating external evidence into vision--language reasoning. Early works such as KAT~\citep{gui-etal-2022-kat}, REVIVE~\citep{lin2022revive}, RA-VQA~\citep{lin2022retrieval}, and MuRAG~\citep{chen-etal-2022-murag} improve performance with retrieved textual knowledge, but largely rely on static queries and text-dominant similarity. More recent frameworks~\citep{yan2024echosight,zhang2024mr,qi2024rora,tian2025core,ling2025mmkb} combine hierarchical or multimodal retrieval, reranking, and reflection strategies, while VisRAG~\citep{yu2024visrag} and M3DocRAG~\citep{cho2025m3docvqa} focus on multimodal document understanding. However, most still follow a static retrieve--rerank--generate pipeline, making cross-modal mismatch and spurious evidence selection common. More recent agentic methods, such as MMSearch-R1~\citep{wu2025mmsearch}, OmniSearch~\citep{li2025benchmarking}, and SenseNova-MARS~\citep{xien2025sensenova}, incorporate query decomposition and tool use, but they still separate visual and textual retrieval, lack explicit mechanisms for resolving cross-modal conflicts, and do not provide auditable reasoning trajectories. In contrast, our approach explicitly parses retrieval intent before retrieval and selects an appropriate structured knowledge space for grounded evidence acquisition, yielding a more interpretable retrieval process for complex multimodal queries.
\section{MM-R2 Framework}
\label{sec:framework}

% \vspace{-3mm}
% \subsection{Task Definition}
% \label{subsec:task_definition}
\textbf{Task Definition.} 
Given a text question $Q$ and an image $I$, mRAG aims to retrieve supporting evidence  from an external corpus   and generate a grounded answer $\hat{y}$.
\textbf{MM-R2 Overview.}
The overview of our
proposed framework is illustrated in Figure ~\ref{fig1}.
We organize the external corpus into a KnowledgeMap $\mathcal{K}=\{u_1,\dots,u_N\}$, where each $u_n$ denotes a semantically coherent retrieval unit (detailed in section \ref{subsec:knowledgemap}). 
We formulate mRAG as a sequential decision-making problem, where an agent iteratively selects what evidence to retrieve given the current state. At each step, the agent operates on a structured state that encodes question intent and visual grounding, and selects $u_n$ to retrieve relevant evidence.

Formally, given a pair of $(I,Q)$, MM-R2 first extracts a structured intent triplet
\begin{equation} \label{eq:intent}
z=(i,r,c)=\Phi(I,Q),
\end{equation}
where $i$ denotes the information need, $r$ denotes the grounded referent in the image, and $c$ represents task-specific constraints. $\Phi$ denotes the fixed Intent-Referent Binding Module, which converts the image-question pair into a structured retrieval state (see section \ref{subsec:IGB}  \emph{Intent-Referent Binding Module}).
% \textcolor{red}{What is $\Phi$?} (see section \ref{subsec:IGB}  \emph{Intent-Referent Binding Module}).

Initially conditioned on the intent $z$ (and subsequently conditioned on the evolving retrieval history $h_t$), the agent $\mathcal{A}$ performs two decisions: a routing action $a_t$ that selects the retrieval unit(s) to search, and a generated  query $q_t$ that retrieves relevant evidence within the selected unit. This yields the retrieved evidence $e_t$.
The retrieval history is then updated as
\begin{equation}
h_{t+1}=(h_t,(a_t,q_t,e_t)).
\end{equation}
After each retrieval step (before reaching a predefined maximum retrieval budget), the agent reasons over the current history to decide whether to continue retrieval or to terminate the retrieval loop and generate the final answer. %. The process stops when the agent outputs an answer or when a predefined maximum retrieval budget is reached.
% After each retrieval step, the agent reasons over the current history to determine whether additional retrieval is needed.
\begin{figure*}[t]
\centering
\includegraphics[width=1\textwidth]{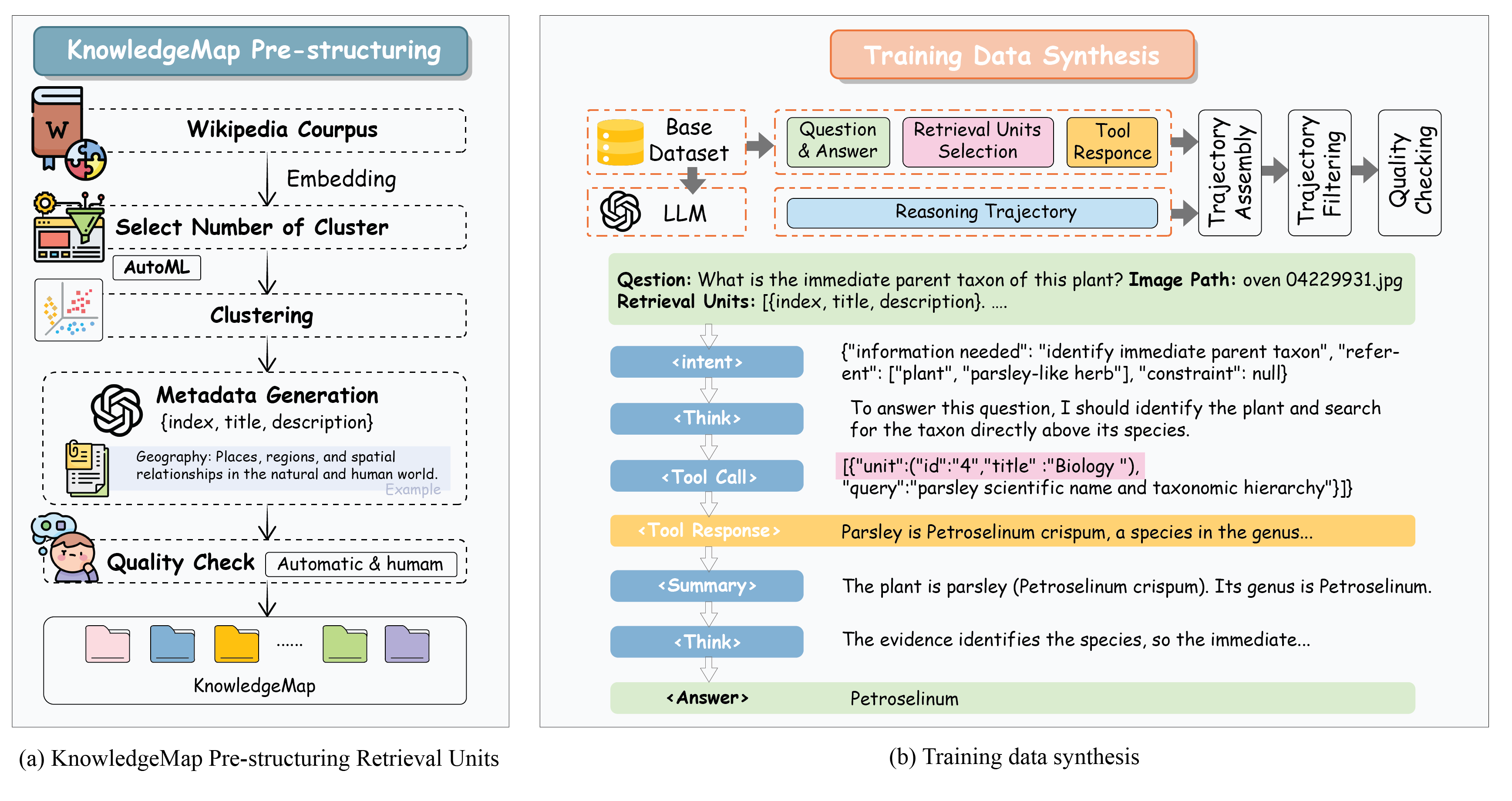}
% \vspace{-12pt}
\caption{\textbf{Offline construction of the KnowledgeMap (a)}:
the corpus is pre-structured into semantically coherent retrieval units through embedding, clustering, metadata generation, and quality checking. %, yielding an interpretable KnowledgeMap.
\textbf{Synthesis of MM-R2-Traj (b)}: given an image-question pair and the KnowledgeMap, a teacher model synthesizes structured multi-step retrieval trajectories, including intent states, reasoning, tool calls, tool responses, summaries, and final answers.
} 
% \vspace{-15pt}
\label{fig2}
\end{figure*}
Finally, the answer is generated by the answer generator (often the same agent $\mathcal{A}$) based on the original multimodal input, the intent, and the retrieval history:
\begin{equation}
\hat{y}=\mathcal{A}(I,Q,z,h_t).
\end{equation}

In the MM-R2 framework, only the agent $\mathcal{A}$ needs to be trained, while  the Intent-Referent Binding $\Phi$ is kept fixed. The training process is detailed in Section \ref{sec:training}.

Unlike conventional mRAG approaches that rely on static similarity scoring or one-shot retrieval, MM-R2  enables adaptive, state-aware evidence selection over multiple steps. 
The agent begins with the inferred intent and iteratively refines both where to search and what evidence to retrieve as additional context is accumulated. This enables the agent to resolve ambiguity and compose evidence in a goal-directed manner, leading to more coherent and effective multimodal reasoning.

%The novelty of our approach lies in (i) elevating retrieval from a passive ranking problem to an active decision-making process, (ii) introducing structured retrieval states that explicitly model intent, grounding, and constraints, and (iii) operating over semantically coherent evidence units rather than isolated items, allowing the agent to reason over meaningful multimodal contexts. This design enables the agent to iteratively refine its retrieval strategy, resolve ambiguity, and compose evidence in a goal-directed manner, leading to more coherent and effective multimodal reasoning. 
% \vspace{-6pt}
\subsection{Intent-Referent Binding Module}
\label{subsec:IGB}% \vspace{-6pt}

This module aims to explicitly bind the question’s information need to its grounded visual referent in the image, yielding a structured intent state that supports subsequent retrieval planning (see Eq. (\ref{eq:intent})).
Given a multimodal query $(I,Q)$, the Intent-Referent Binding Module $\Phi$ maps the image-question pair to a structured triplet $(i,r,c)$. Here, the information need $i$ captures what knowledge is being asked for, such as taxonomy, location, historical background, or functional property; the grounded visual referent $r$ identifies the image-grounded object, entity, region, or scene element that the query refers to; and the task-specific constraints $c$ encode additional retrieval conditions such as answer type, granularity, temporal scope, geographic scope, or other disambiguating cues. In practice, $\Phi$ is implemented as a fixed prompting-based module over a multimodal language model that outputs these three components in natural language. For example, given an image of a plant and the question ``What is the closest parent taxonomy of this plant?'', $\Phi$ produces an intent state in which $i$ corresponds to the parent taxonomy being queried, $r$ refers to the plant shown in the image, and $c$ specifies that retrieval should focus on biological classification at the nearest higher taxonomic level. Implementation details are provided in Appendix~\ref{app:intent_binding_prompt}.

Representing the intent as a triplet facilitates retrieval by making both the information need and task constraints explicit. This helps identify the most appropriate retrieval unit and enables the generation of within-unit queries that are more specific, less ambiguous, and better aligned with the original multimodal input.
In addition, exposing such information makes retrieval process more interpretable and verifiable.
%Importantly, this representation is produced once from the original multimodal query and then used as a fixed conditioning signal during retrieval, rather than being recursively redefined as a separate symbolic state at each step. In practice, $\Phi$ can be implemented either as a dedicated multimodal encoder or as an internal stage of the same agent, where the agent first predicts the triplet from $(I,Q)$ and then conditions on it for downstream retrieval and answer generation.

%By making intent, grounding, and constraints explicit, intent induction provides a stable foundation for sequential retrieval. This is particularly important for multi-hop or visually ambiguous questions, where relevant evidence cannot be reliably obtained through a single global retrieval step from raw multimodal input.

% \begin{figure*}[t]
% \centering
% \includegraphics[width=0.8\textwidth]{fig/2.png}
% % \vspace{-12pt}
% \caption{\textbf{Offline construction of the KnowledgeMap (a)}:
% the corpus is pre-structured into semantically coherent retrieval units through embedding, clustering, metadata generation, and quality checking. %, yielding an interpretable KnowledgeMap.
% \textbf{Synthesis of MM-R2-Traj (b)}: given an image-question pair and the KnowledgeMap, a teacher model synthesizes structured multi-step retrieval trajectories, including intent states, reasoning, tool calls, tool responses, summaries, and final answers.
% } 
% % \vspace{-15pt}
% \label{fig2}
% \end{figure*}

% \vspace{-2mm}
\subsection{KnowledgeMap: Structuring the Retrieval Units}
\label{subsec:knowledgemap}

We build the KnowledgeMap offline through a pre-structuring pipeline as shown in the right part of Figure~\ref{fig2}. Starting from the InfoSeek~\citep{chen2023can} knowledge base containing approximately 6M Wikipedia-derived passages, we first encode each passage into a dense embedding space. We then perform automated model selection over HDBSCAN~\citep{mcinnes2017hdbscan} configurations to identify a clustering setting that yields semantically coherent and reasonably balanced units. HDBSCAN is particularly suitable here because it can discover clusters of varying density without fixing the number of clusters in advance, which is important for a large and semantically diverse corpus. Each resulting cluster is treated as a candidate retrieval unit. To make these units interpretable for routing,  we use a large language model to generate lightweight metadata, including a short title and a natural-language description summarizing its dominant content. We further apply a quality-checking stage to refine noisy, imbalanced, or semantically incoherent units.

Each passage is assigned to a primary unit,  forming an approximate partition of the corpus. We emphasize that this partition is not intended to be uniquely correct or semantically exhaustive. Rather, the KnowledgeMap functions as an approximate routing scaffold that exposes a more structured and interpretable interface over an otherwise flat search space. Even when unit boundaries are imperfect or a query spans multiple units, this organization remains valuable: it narrows the candidate space, reduces competition from irrelevant semantic regions, and provides a practical basis for explicit \emph{where-to-search} decisions.

Note that the KnowledgeMap does not replace passage-level retrieval; rather, it structures the decision of where retrieval should be performed, while actual evidence retrieval remains a passage-level operation within the selected unit(s).

%During planning, the agent reasons over this structured interface rather than over the full corpus, making routing an explicit and observable decision. In this sense, the intent induction module determines what evidence would count as relevant, while the KnowledgeMap determines where such evidence is most likely to be found. The KnowledgeMap does not replace passage-level retrieval; rather, it structures the decision of where retrieval should be performed, while fine-grained evidence ranking is handled by a fixed retrieval backend wrapped as a retrieval tool.
% \vspace{-2mm}
\subsection{Routing and Query Generation Policy}
%Agentic Retrieval Policy}
\label{subsec:policy}

We next describe how the agent determines the routing action $a_t$ and generates the retrieval query $q_t$ at each step. The agent $\mathcal{A}$ is implemented as a multimodal language model that performs retrieval planning conditioned on the multimodal query, the structured intent state, and the current retrieval history.

At step $t$, the agent takes as input the image $I$, the question $Q$, the intent triplet $z=(i,r,c)$, the current retrieval history $h_t$, and a compact representation of the KnowledgeMap $\mathcal{K}$. It first predicts a routing action $a_t$ that selects one or more retrieval units from $\mathcal{K}$, and then generates a grounded retrieval query $q_t$ conditioned on $(I,Q,z,h_t,a_t)$. In this way, routing determines where to search, while query generation determines what to search for within the selected region.

The action $(a_t,q_t)$ is executed by an external retrieval tool restricted to the selected retrieval unit(s), yielding retrieved evidence $e_t=T_{\mathrm{ret}}(q_t,a_t)$. In our implementation, $T_{\mathrm{ret}}$ is a passage-level retriever that searches only within the passages assigned to the selected unit(s). The retrieval history is then updated as $h_{t+1}=(h_t,(a_t,q_t,e_t))$. The updated history is appended to the interaction context and used by the agent in subsequent steps to refine routing, formulate new queries, and decide whether additional retrieval is necessary. The exact prompt templates used for routing and query generation are provided in Appendix~\ref{app:routing_prompt}.

%\input{sec/_sec_4_TrainingData}
% \vspace{-3mm}
\section{Agent Training}
\label{sec:training}
% \vspace{-3mm}

\begin{figure*}[t]
\centering
\includegraphics[width=1\textwidth]{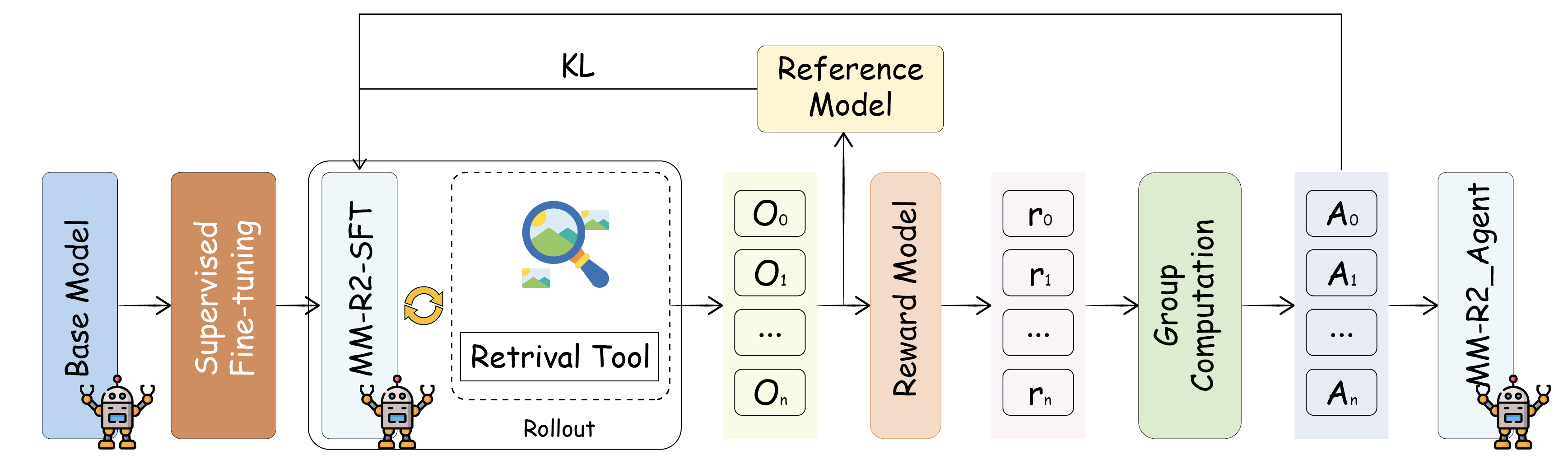}
% \vspace{-8pt}
\caption{\textbf{Overview of the two-stage training strategy: SFT+GRPO}
} 
% \vspace{-10pt}
\label{fig3}
\end{figure*}

\subsection{Training Data Synthesis}
\label{sec:data}% \vspace{-6pt}

To teach the agent to perform multi-step retrieval planning, we construct supervised retrieval trajectories for multimodal question answering, denoted as MM-R2-Traj. The right part of Figure~\ref{fig2} illustrates the overall synthesis pipeline. Each trajectory corresponds to a single image-question pair and records a sequence of retrieval steps. At an abstract level, a trajectory is represented as $((a_1,q_1,e_1), \dots, (a_T,q_T,e_T))$, where $a_t$ denotes the retrieval unit selected at step $t$, $q_t$ denotes the grounded retrieval query issued at that step, and $e_t$ denotes the evidence returned by the retriever.

To synthesize such trajectories at scale, we use the training splits of InfoSeek~\citep{chen2023pretrainedvisionlanguagemodels} and Encyclopedic VQA~\citep{mensink2023encyclopedic}, which contain large-scale multimodal question-answer pairs. For each image-question pair $(I,Q)$, we first derive the structured intent state $z=(i,r,c)$ using the Intent-Referent Binding Module. Based on $(I,Q,z)$ and the KnowledgeMap, we then construct candidate retrieval traces and use a teacher multimodal reasoning model to normalize them into structured retrieval trajectories, including retrieval-unit selection, retrieval-query generation, evidence inspection, and final answer derivation. The exact teacher model and prompt templates are provided in Appendix~\ref{app:trajectory_synthesis}.

Rather than directly using the teacher's full free-form reasoning traces, we convert each example into the same structured textual format used by our agent. Concretely, each retrieval step is serialized with explicit XML-style tags. The tag \texttt{<think>} records the model's intermediate reasoning before or after retrieval. The tag \texttt{<tool call>} contains a structured retrieval action, including both the selected retrieval unit and the issued query; in particular, the unit id corresponds to $a_t$, while the query field corresponds to $q_t$. The evidence returned by the retriever is wrapped by \texttt{<tool response>}, corresponding to $e_t$. In addition, \texttt{<summary>} records a concise factual condensation of the retrieved evidence for use in later steps, and \texttt{<answer>} marks the final answer. In this way, the trajectory format remains directly grounded in the underlying retrieval process, while preserving the intermediate reasoning and evidence integration behavior that the model must learn.

To ensure data quality, we apply both automatic filtering and manual inspection. Automatically, we remove trajectories with incorrect final answers, incomplete retrieval steps, inconsistent summaries, or degenerate reasoning patterns. Here, the reference final answer is used only for post-hoc verification and filtering, rather than for authoring the reasoning trace itself. We further conduct human inspection on randomly sampled trajectories to verify retrieval-unit selection quality, query grounding, rationale faithfulness, and evidence support. After filtering, the final MM-R2-Traj dataset contains 900K trajectories, of which 22K involve two or three retrieval steps. The average serialized trajectory length is 2949 tokens. More details and examples are provided in Appendix~\ref{app:trajectory_synthesis}.

%Our goal is to train MM-R2 as an agent capable of performing structured multi-step retrieval over the KnowledgeMap and producing grounded answers through iterative interaction with the retrieval environment. Starting from the synthetic
% \vspace{-6pt}
\subsection{Stage 1: Supervised Fine-Tuning (SFT)}
% \vspace{-6pt}
With the trajectory dataset MM-R2-Traj, we adopt a two-stage training strategy, as illustrated in Figure~\ref{fig3}. In the first stage, we perform supervised fine-tuning (SFT) as a cold start, enabling the agent to learn the basic structure of multi-step retrieval behavior, including retrieval-unit selection, grounded query formulation, evidence summarization, and answer generation. In the second stage, we further optimize the agent with Group Relative Policy Optimization (GRPO), so that it can refine its retrieval policy under task-level rewards. This two-stage design combines the stability of imitation learning with the adaptability of reinforcement learning. 
%In the first stage, we perform supervised fine-tuning on the base vision-language model using MM-R2-Traj. This cold-start stage equips the model with the foundational ability to produce structured multi-step retrieval trajectories under our agent framework.

%Each training trajectory is represented as a structured multi-turn interaction sequence, in which the model alternates between internal reasoning, retrieval actions, intermediate summarization, and final answer generation. 

%The objective of this stage is to equip the model with the ability to generate well-formed multi-step retrieval trajectories, including reasoning, retrieval-unit selection, grounded query generation, evidence summarization, and answer generation. Since the content inside \texttt{<tool\_response>}...\texttt{</tool\_response>} is produced by the external retrieval environment rather than by the model itself, these observation tokens are masked out from the loss computation~\citep{chen2023fireact}. Therefore, 

In the trajectories,  the content enclosed by \texttt{<tool\_response>}...\texttt{</tool\_response>} is produced by the external retrieval environment, and is therefore not treated as a learning target. Following prior work~\citep{chen2023fireact}, these observation tokens are masked out during loss computation. Accordingly,  
the SFT objective maximizes the log-likelihood only over model-generated segments, including reasoning, summarization, action, and answer tokens:\texttt{<intent>}, \texttt{<think>}, \texttt{<summary>}, \texttt{<tool\_call>}, and \texttt{<answer>}. 
%In this way, the model learns how to select appropriate KnowledgeMap branches, generate grounded retrieval queries for the selected branches, integrate retrieved evidence, and produce the final answer, while treating tool outputs as external context.
After this stage, the agent can generate well-formed multi-step retrieval trajectories and serves as the initialization for the subsequent reinforcement learning stage.

\subsection{Stage 2: Group Relative Policy Optimization (GRPO)}
% \vspace{-6pt}
Starting from the SFT-initialized policy, we further optimize the agent with GRPO ~\citep{shao2024deepseekmath} to improve retrieval planning under task-level feedback. For each query, we sample a group of rollouts by interacting with the retrieval environment, where each rollout contains the model-generated reasoning and retrieval actions together with the corresponding tool responses. GRPO updates the policy by comparatively favoring rollouts with better overall outcomes. We omit the standard GRPO objective here for brevity and provide the full optimization details in Appendix~\ref{app:two_stage_training}.

\noindent
\textbf{Reward
design.}
Our GRPO training uses a rollout-level reward that jointly captures routing quality and final answer correctness:
\begin{equation}
R_i
=
\lambda_{\mathrm{route}} R_i^{\mathrm{route}}
+
\lambda_{\mathrm{ans}} R_i^{\mathrm{ans}}.
\end{equation}
Here, $R_i^{\mathrm{route}}$ evaluates the quality of retrieval routing throughout rollout $i$, while $R_i^{\mathrm{ans}}$ evaluates whether the final answer is correct.

Specifically, at each retrieval step, the agent selects a branch of the KnowledgeMap and issues a grounded query within that branch. We assign a step-level routing score $r_{i,t}^{\mathrm{route}}$ according to whether the selected branch at step $t$ is appropriate, and then aggregate these scores across the trajectory:
\begin{equation}
R_i^{\mathrm{route}}
=
\frac{1}{T_i}\sum_{t=1}^{T_i} r_{i,t}^{\mathrm{route}},
\end{equation}
where $T_i$ is the number of retrieval steps in rollout $i$. The answer reward $R_i^{\mathrm{ans}}$ is computed by comparing the final predicted answer with the reference answer.

This reward design encourages the model to improve both step-wise retrieval behavior and final task success. During rollout generation, the model sequentially produces reasoning and retrieval actions, receives tool feedback from the environment, and updates subsequent decisions based on the retrieved evidence. Each rollout terminates when the model outputs a final answer or reaches a predefined maximum retrieval budget.

\section{Evaluation Setup}
% \vspace{-1mm}
\subsection{Two-Axis Evaluation for Agentic RAG}
% \vspace{-1mm}
Because MM-R2 explicitly exposes intermediate retrieval decisions, we evaluate it along two complementary axes: task-level effectiveness and process-level verifiability. 
The first measures final answer correctness; the second assesses  whether the system makes explicit where it searches, what evidence it uses, and whether the exposed trace can genuinely justify the answer. Additional details of metric construction and evaluation protocols are provided in Appendix~\ref{evl}.
% \vspace{-1mm}

\noindent\textbf{Axis 1: Task-level Effectiveness.}
We evaluate both final answer quality and retrieval quality. For the former, we report standard QA metrics, including Accuracy and F1 when applicable. For the latter, we report Recall@k, which measures whether the gold evidence or target retrieval unit appears among the top-k retrieved results.
% \vspace{-1mm}

\noindent\textbf{Axis 2: Process-level Verifiability.}
We evaluate whether the exposed retrieval trajectory reflects the intended behavior of MM-R2.

\noindent\textbf{- LLM-as-a-Judge (LJ).}
A strong LLM scores (i) trace--answer consistency and (ii) evidence support given the query, the exposed trajectory, and the retrieved evidence.
% \vspace{-1pt}

\noindent\textbf{- Route Recall@1.}
This metric measures whether the agent routes to the correct KnowledgeMap region. With $U_q^\star$ denoting the gold unit(s) and $\hat{U}_q$ the predicted route, RouteRecall@1 is defined as \(\frac{1}{|Q|}\sum_{q\in Q}\mathbf{1}[\hat{U}_q \cap U_q^\star \neq \emptyset]\).
% \vspace{-1pt}

\noindent\textbf{- Explanation Sparsity.}
This metric measures the conciseness of the exposed evidence summary: \(\frac{1}{|Q|}\sum_{q\in Q}|E_q|\), where lower values are better.
% \vspace{-1pt}

\noindent\textbf{- Fidelity.}
This metric measures whether the final prediction is reproducible from the exposed evidence summary: \(\frac{\mathrm{Acc}_{\mathrm{evid}}}{\mathrm{Acc}_{\mathrm{full}}}\). Higher values indicate that the revealed evidence better accounts for the model's final decision.

\section{Experiments Results}
% \vspace{-6pt}

% Detailed experimental settings, including datasets, implementation details, and baselines, are provided in Appendix~\ref{setup}.

\subsection{Experimental Setup }
\label{setup}

\noindent
\textbf{Datasets.} 
We conduct experiments on two large-scale knowledge-based VQA benchmarks, InfoSeek~\citep{chen2023pretrainedvisionlanguagemodels} and Encyclopedic VQA~\citep{mensink2023encyclopedic}. Both benchmarks require grounding image--text queries with external encyclopedic knowledge.InfoSeek contains 1.3M image--question--answer triplets linked to a large-scale knowledge base. Following prior work, we evaluate on the official validation split, which contains 73K examples and is further divided into the Unseen-Entity (Unseen-E) and Unseen-Question (Unseen-Q) subsets. Encyclopedic VQA~\citep{mensink2023encyclopedic} is a knowledge-intensive visual question answering benchmark that focuses on encyclopedic reasoning over fine-grained visual entities. It contains 221K unique questions and serves as a complementary evaluation benchmark to InfoSeek. For training, we use MM-R2-Traj, a large-scale trajectory dataset of agentic retrieval processes. MM-R2-Traj contains 900K high-quality trajectories, of which 860K are used for training and 40K are used for validation.

\noindent
\textbf{Implementation Details.} 
MM-R2 is instantiated with Qwen2-VL-7B~\citep{wang2024qwen2} as the backbone multimodal language model for the trainable agent. Retrieval is performed by an external dense retriever, with the search space restricted to the selected KnowledgeMap unit(s). For process-level evaluation, the LLM-as-a-Judge is implemented using GPT-4o~\citep{hurst2024gpt}. We use the TRL library~\citep{vonwerra2022trl} for both supervised fine-tuning (SFT) and reinforcement learning (RL). For SFT, we train for 5{,}000 steps with a batch size of 8 and a learning rate of $5\times 10^{-6}$. For RL, we train for 90 steps with a batch size of 16, a learning rate of $5\times 10^{-7}$, a sampling temperature of 1.0, and top-$p$ of 1.0. The rollout-level reward is defined as
\[
R_i=\lambda_{\mathrm{route}}R_i^{\mathrm{route}}+\lambda_{\mathrm{ans}}R_i^{\mathrm{ans}},
\]
where $\lambda_{\mathrm{route}}=0.1$ and $\lambda_{\mathrm{ans}}=1.0$. We also enable DeepSpeed ZeRO-3~\citep{rajbhandari2020zeromemoryoptimizationstraining} for memory-efficient distributed training. Our model is implemented in PyTorch and trained on 8$\times$NVIDIA H100 GPUs.

\noindent
\textbf{Baselines.}
We compare MM-R2 against a broad spectrum of zero-shot, retrieval-augmented, and fine-tuned vision--language baselines, as summarized in Table~\ref{tab:main_results}. 
(1) \textbf{BLIP-2}~\citep{li2023blip} and 
(2) \textbf{InstructBLIP}~\citep{dai2023instructblip} are representative zero-shot vision--language models that answer questions without external retrieval.
(3) \textbf{LLaVA-1.5}~\citep{liu2023visual} is a Vicuna-7B based vision--language model used as a strong open-source parametric baseline without an external knowledge base.
(4) \textbf{GPT-4V}~\citep{achiam2023gpt} serves as a strong proprietary multimodal baseline without explicit retrieval.
(5) \textbf{Qwen2-VL-Param}~\citep{wang2024qwen2} is a Qwen2-VL-7B baseline that answers questions purely from its parametric knowledge.
(6) \textbf{Qwen2-VL (KnowledgeMap)}~\citep{wang2024qwen2} uses the same backbone but performs retrieval over our structured KnowledgeMap, providing a zero-shot retrieval baseline that isolates the effect of the retrieval interface without agent training.
(7) \textbf{CoMEM}~\citep{wu2025towards} adds a continuous-memory module that compresses retrieved evidence into a small set of memory vectors; we report both Qwen2-VL-7B and Qwen2.5-VL-7B variants.
Among fine-tuned retrieval-augmented baselines, 
(8) \textbf{Wiki-LLaVA}~\citep{caffagni2024wiki} extends LLaVA with Wikipedia retrieval and instruction tuning, yielding a strong flat mRAG baseline over encyclopedic knowledge. 
(9) \textbf{RoRA-VLM}~\citep{qi2024rora} combines Wikipedia and web search and performs multi-hop reasoning over mixed textual evidence. 
(10) \textbf{EchoSight}~\citep{yan2024echosight} is a LLaMA3-8B based encyclopedic VQA system that jointly learns retrieval and answer prediction on Wikipedia. 
(11) \textbf{ReflectiVA}~\citep{cocchi2025augmenting} introduces self-reflective tokens to predict whether retrieval is needed and to assess the relevance of retrieved evidence. 
(12) \textbf{LLaVA-mR$^2$AG}~\citep{zhang2024mr} equips LLaVA with a multi-round retrieve--reason pipeline for iterative evidence refinement. 
(13) \textbf{Wiki-PRF-7B} and \textbf{Wiki-R1-7B}~\citep{hong2025knowledge} are Qwen2.5-VL-7B based RAG systems that optimize retrieval and reasoning with process supervision and reinforcement learning. 
(14) \textbf{CoRe-MMRAG}~\citep{tian2025core} is a supervised cross-modal mRAG framework. 
For fairness, we report the numbers from the original papers when available; note that on Enc-VQA, \textbf{LLaVA-mR$^2$AG} uses Google Lens retrieval, as indicated in Table~\ref{tab:main_results}.

\definecolor{lightgrayrow}{RGB}{225,225,225}
\definecolor{lightpurplerow}{RGB}{228,212,245}

% in main text
\begin{table*}[t]
\centering
\renewcommand{\arraystretch}{1.08}
\resizebox{1\textwidth}{!}{
\begin{tabular}{lcccccc}
\toprule
\multirow{2}{*}{\bf Method} & \multirow{2}{*}{\bf Model} & \multirow{2}{*}{\bf KB} & \multirow{2}{*}{\bf Enc-VQA} & \multicolumn{3}{c}{\bf InfoSeek} \\
& & & & \bf Unseen-Q & \bf Unseen-E & \bf All \\
\cmidrule(lr){5-7}
\midrule

\rowcolor{lightgrayrow}
\multicolumn{7}{l}{\bf Zero-shot Models} \\
BLIP-2~\citep{li2023blip} & Flan-T5XL & - & 12.6 & 12.7 & 12.3 & 12.5 \\
InstructBLIP~\citep{dai2023instructblip} & Flan-T5XL & - & 11.9 & 8.9 & 7.4 & 8.1 \\
LLaVA-1.5~\citep{liu2023visual} & Vicuna-7B & - & 16.3 & 13.0 & 10.3 & 12.2 \\
GPT-4V~\citep{achiam2023gpt} & - & - & 26.9 & 15.0 & 14.3 & 14.6 \\
Qwen2-VL-Param~\citep{wang2024qwen2} & Qwen2-7B & - & 12.7 & 23.1 & 21.8 & 22.1 \\
Qwen2-VL (KnowledgeMap)~\citep{wang2024qwen2} & Qwen2-7B & Wiki & 7.6 & 13.3 & 11.1 & 11.2 \\
CoMEM (Qwen2-VL)~\citep{wu2025towards} & Qwen2-7B & Wiki & - & 32.6 & 33.1 & 32.9 \\
CoMEM (Qwen2.5-VL)~\citep{wu2025towards} & Qwen2.5-7B & Wiki & - & 32.8 & 28.5 & 30.7 \\
\midrule

\rowcolor{lightgrayrow}
\multicolumn{7}{l}{\bf Fine-tuned Models} \\
Wiki-LLaVA~\citep{caffagni2024wiki} & Vicuna-7B & Wiki & 17.7 & 30.1 & 27.8 & 28.9 \\
RoRA-VLM~\citep{qi2024rora} & Vicuna-7B & Wiki+Web & 20.3 & 27.3 & 25.1 & 26.9 \\
EchoSight~\citep{yan2024echosight} & LLaMA3-8B & Wiki & 19.4 & - & - & 27.7 \\
ReflectiVA~\citep{cocchi2025augmenting} & LLaMA-3.1-8B & Wiki & 35.5 & 40.4 & 39.8 & 40.1 \\
LLaVA-mR$^2$AG~\citep{zhang2024mr} & Vicuna-7B & Wiki & 55.1* & 39.1 & 39.7 & 39.4 \\
Wiki-PRF-7B (SFT)~\citep{hong2025knowledge} & Qwen2.5-7B & Wiki & - & 41.5 & 41.9 & 41.8 \\
Wiki-PRF-7B (RL)~\citep{hong2025knowledge} & Qwen2.5-7B & Wiki & - & 46.6 & 46.2 & 46.3 \\
Wiki-R1-7B~\citep{hong2025knowledge} & Qwen2.5-7B & Wiki & - & 47.8 & 42.3 & 44.1 \\
CoRe-MMRAG~\citep{tian2025core} & Qwen2-7B & Wiki & 27.2 & 45.2 & 46.9 & 46.5 \\
\rowcolor{lightpurplerow}
\textbf{MM-R2 (ours)} & Qwen2-7B & Wiki & \textbf{39.4} & \textbf{54.1} & \textbf{56.0} & \textbf{54.3} \\
\bottomrule
\end{tabular}
}
\caption{
\textbf{Main results (\%) on Enc-VQA and InfoSeek with external knowledge.}}
\label{tab:main_results}
\end{table*}

% \vspace{-6pt}
\subsection{Task-level Effectiveness}
% \vspace{-6pt}
\begin{wraptable}{r}{0.45\columnwidth}
% \vspace{-15pt}
\centering
\small
\setlength{\tabcolsep}{3pt}
\begin{tabular}{lccc}
\toprule
\textbf{Method} & \textbf{Top-1} & \textbf{Top-2} & \textbf{Top-5} \\
\midrule
Qwen2 (+KM) & 11.2 & 13.1 & 15.5 \\
CoRe-MMRAG 1-Stage & 39.6 & 40.4 & 40.9 \\
CoRe-MMRAG & 40.7 & 42.9 & 42.9 \\
Wiki-PRF-7B & 38.9 & -- & 39.5 \\
MM-R2 (ours) & \textbf{47.9} & \textbf{51.0} & \textbf{54.3} \\
\bottomrule
\end{tabular}
% \vspace{-4pt}
\caption{\textbf{Overall accuracy (\%) on InfoSeek under different retrieval budgets.}}
% \vspace{-8pt}
\label{tab:infoseek_topk_compare}
\end{wraptable}
Table~\ref{tab:main_results} reports the main task-level results on Enc-VQA and InfoSeek. %Existing zero-shot baselines remain relatively weak on both benchmarks. Purely parametric models such as BLIP-2, InstructBLIP, LLaVA-1.5, GPT-4V, and Qwen2-VL-Param achieve limited performance, while zero-shot retrieval-based variants such as Qwen2-VL (KnowledgeMap) and CoMEM improve retrieval grounding but still lag behind strong fine-tuned systems on InfoSeek. Fine-tuned baselines further narrow the gap, yet most methods remain below 50\% overall accuracy. Among prior methods, CoRe-MMRAG is the strongest overall baseline on InfoSeek, achieving 46.5\% overall accuracy, while Wiki-R1-7B attains the best prior result on the Unseen-Q split at 47.8\%.
MM-R2 achieves 54.1\%, 56.0\%, and 54.3\% on InfoSeek Unseen-Q, Unseen-E, and All, improving over the strongest prior baselines by 6.3, 9.1, and 7.8 points, respectively. On Enc-VQA, MM-R2 reaches 39.4\%, outperforming most baselines. Although lower than the 55.1\% reported by LLaVA-mR$^2$AG, that result relies on Google Lens retrieval, whereas MM-R2 uses a dense retriever over our KnowledgeMap-based pipeline. Overall, these gains suggest that MM-R2 benefits from explicitly modeling both \emph{what} to retrieve and \emph{where} to search.
Table~\ref{tab:infoseek_topk_compare} further %compares InfoSeek overall accuracy under different retrieval budgets. 
confirms that MM-R2 consistently outperforms prior methods across all settings, achieving 47.9\%, 51.0\%, and 54.3\% under Top-1, Top-2, and Top-5 retrieval budgets, respectively. %Compared with the strongest prior results under the same budgets, MM-R2 improves performance by 7.2 points at Top-1, 8.1 points at Top-2, and 11.4 points at Top-5. 
Notably, the gains become larger as the retrieval budget increases, suggesting that MM-R2 not only makes stronger first-step routing decisions, but also benefits more effectively from additional retrieved evidence when multiple candidates are available.

% \begin{table}[t]
% \centering
% \small
% \resizebox{0.5\columnwidth}{!}{
% \begin{tabular}{lccc}
% \toprule
% \textbf{Method} & \textbf{Top-1} & \textbf{Top-2} & \textbf{Top-5} \\
% \midrule
% Qwen 2 (+KnowledgeMap)  & 11.2 & 13.1 & 15.5 \\
% CoRe-MMRAG 1-Stage & 39.6 & 40.4 & 40.9 \\
% CoRe-MMRAG & 40.7 & 42.9 & 42.9 \\
% Wiki-PRF-7B & 38.9 & -- & 39.5 \\
% MM-R2 (ours) & \textbf{47.9} & \textbf{51.0} & \textbf{54.3} \\
% \bottomrule
% \end{tabular}
% }
% \caption{\textbf{Comparison of overall accuracy (\%) on InfoSeek under different retrieval budgets.} }
% \label{tab:infoseek_topk_compare}
% \end{table}

\begin{figure*}[h!]
\centering
\includegraphics[width=1\textwidth]{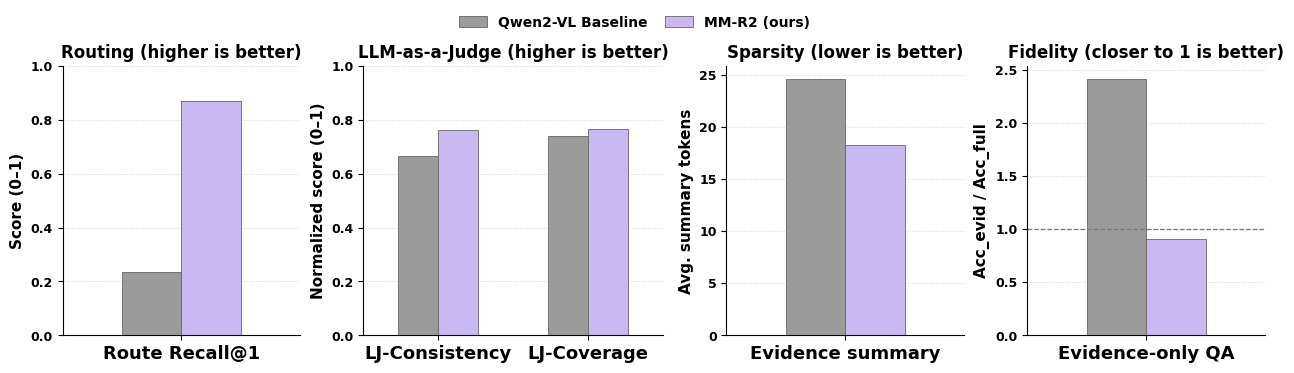}
% \vspace{-5pt}
\caption{
\textbf{Process-level verifiability on InfoSeek.}
MM-R2 improves route quality and judge-based trace quality over the Qwen2-VL baseline, while producing more concise evidence summaries and fidelity values much closer to 1.
}
% \vspace{-6pt}
\label{fig4}
\end{figure*}

% \vspace{-6pt}
\subsection{Process-level Effectiveness}% \vspace{-6pt}

As shown in Figure~\ref{fig4}, MM-R2 substantially improves the quality and verifiability of the retrieval process. 
First, MM-R2 achieves a large gain in \emph{routing accuracy}, with Route Recall@1 improving from 0.2339 to 0.7400, suggesting  that the model is much more likely to access the correct KnowledgeMap branch rather than exploring irrelevant regions of the search space. 
%We attribute this improvement to two key components of MM-R2: the Intent-Referent Binding module, which explicitly resolves \emph{what} the query is asking about and \emph{which} visual referent it concerns, and the structured KnowledgeMap, which turns retrieval-space selection into an explicit routing decision rather than leaving it implicit in flat similarity matching. 
%Together, these components make the retrieval workflow more targeted and substantially reduce early-stage search drift.
Second,  MM-R2  produced traces that are not only more coherent with the final answer but also better supported by the retrieved evidence, as indicated by the improved \emph{LLM-as-a-Judge consistency} and \emph{coverage}, from 0.6677 to 0.7620 and from 0.7414 to 0.7660, respectively. 
%Beyond routing, MM-R2 also improves the quality of the exposed reasoning traces. 
%Both LLM-as-a-Judge consistency and coverage increase, from 0.6677 to 0.7620 and from 0.7414 to 0.7660, respectively, indicating that the produced traces are not only more coherent with the final answer but also better supported by the retrieved evidence. 
At the same time, \emph{Explanation Sparsity} decreases from 24.5840 to 18.2820 tokens, showing that MM-R2 produces more concise explanations while preserving or improving evidential support. 
%This pattern is consistent with the design of MM-R2: once retrieval intent is better grounded and routing is more accurate, the model can retrieve more relevant evidence earlier, reducing the need for verbose or noisy explanations.
Last, the \emph{Fidelity} further highlights the difference between MM-R2 and the baseline. 
%The Qwen2-VL baseline yields a fidelity score of 2.4, suggesting a substantial mismatch between the exposed evidence summary and the behavior of the full pipeline.  In contrast, 
MM-R2 attains a fidelity of 0.9011, which is much closer to 1 than baselines and therefore indicates that the final prediction is largely reproducible from the revealed evidence. 
%This result suggests that MM-R2 not only improves final-task accuracy, but also makes the reasoning process itself more faithful and inspectable. 
Overall, these process-level results show that the proposed workflow is effective not only because it retrieves better evidence, but also because it induces a more transparent, compact, and trustworthy reasoning trajectory.
\begin{table*}[t]
\centering

\begin{subtable}[t]{0.48\textwidth}
\centering
\footnotesize
\resizebox{\linewidth}{!}{
\begin{tabular}{lccc}
\toprule
Model & \multicolumn{3}{c}{Acc.} \\
\cmidrule{2-4}
& Unseen-Q & Unseen-E & All \\
\midrule
\textbf{Qwen2-VL} & 13.3 & 11.1 & 11.2 \\
\textbf{MM-R2 (+SFT)} & 53.6 & 55.1 & 53.8 \\
\textbf{MM-R2 (+SFT+RL)} & \textbf{54.1} & \textbf{56.0} & \textbf{54.3} \\
\bottomrule
\end{tabular}
}
\caption{Training ablation on InfoSeek with three split.}
\label{tab:ablation_stage}
\end{subtable}
\hspace{0.02\textwidth}
\begin{subtable}[t]{0.48\textwidth}
\centering
\footnotesize
\resizebox{0.9\linewidth}{!}{
\begin{tabular}{cccc}
\toprule
 Intent  &  KM Routing  &  Agentic Loop &  Acc. \\
\midrule
\xmark & \xmark & \xmark & 22.1 \\
\xmark & \cmark & \xmark & 30.2 \\
\cmark & \cmark & \xmark & 38.7 \\
\cmark & \cmark & \cmark & \textbf{54.3} \\
\bottomrule
\end{tabular}
}
\caption{Module ablation of key MM-R2 components.}
\label{tab:ablation_module}
\end{subtable}% \vspace{-0.1in}
\caption{\textbf{Ablation results on InfoSeek.} Left: training ablation. Right: module ablation.}% SFT provides the major performance gain, while RL and the full MM-R2 workflow further improve accuracy.}
% \caption{\textbf{Ablation results on InfoSeek.} (a) Supervised fine-tuning provides the major performance gain, while RL further improves the SFT-initialized model. (b) Incrementally enabling intent grounding, KnowledgeMap-based routing, and the agentic retrieval loop leads to consistent accuracy improvements.}
\label{tab3}
\end{table*}

% \vspace{-6pt}
\subsection{Ablation Study}% \vspace{-6pt}

We further conduct ablation studies on InfoSeek to examine % the contribution of the two-stage training strategy and 
the effectiveness of key  components and two-stage training strategy, as shown in Table~\ref{tab3}.

\noindent\textbf{Training ablation.}
%The left part of Table~\ref{tab3} evaluates the contribution of supervised fine-tuning (SFT) and reinforcement learning (RL). 
SFT provides the major performance gain: starting from the base Qwen2-VL model, accuracy rises from 13.3\%, 11.1\%, and 11.2\% to 53.6\%, 55.1\%, and 53.8\% on the Unseen-Q, Unseen-E, and All splits, respectively. %This result shows that SFT is crucial for teaching the model the basic structure of the MM-R2 workflow, including following a structured agentic retrieval format, selecting appropriate KnowledgeMap routes, generating grounded retrieval queries, and integrating retrieved evidence into answer generation. Building on this strong initialization, 
RL further improves performance to 54.1\%, 56.0\%, and 54.3\%. Although the gains over SFT are smaller, they are consistent across all splits, indicating that RL does provide meaningful refinement. % rather than merely noisy optimization. %In our framework, this suggests that the two-stage training strategy has a clear division of roles: SFT teaches the model how to perform the intended retrieval procedure, while RL further sharpens routing and evidence-use decisions under task-level feedback.

\noindent\textbf{Module ablation.} The ablation shows a clear cumulative benefit from each component of MM-R2. Starting from 22.1\% accuracy without intent binding, KnowledgeMap routing, or agentic retrieval, performance rises to 30.2\% with routing alone, 38.7\% after adding intent binding, and 54.3\% with the full agentic retrieval loop. This confirms that MM-R2’s gains come from combining explicit intent grounding, structured retrieval-space selection, and iterative evidence acquisition.
%The right part of Table~\ref{tab3} evaluates the incremental contribution of the key MM-R2 components. Starting from the weakest setting without explicit intent binding, KnowledgeMap routing, or the agentic retrieval loop, the model achieves 22.1\% accuracy. Enabling KnowledgeMap-based routing alone improves performance to 30.2\%, showing that explicitly constraining \emph{where} to search already provides a clear benefit over unguided retrieval. Adding intent binding further boosts accuracy to 38.7\%, indicating that explicitly modeling \emph{what} to retrieve and grounding the query to the correct visual referent substantially improves route selection and evidence acquisition. Finally, enabling the full agentic retrieval loop leads to 54.3\% accuracy, yielding the largest overall gain. This suggests that iterative reason--retrieve interaction is critical for composing evidence across steps, rather than relying on a single-shot retrieval decision. Overall, the module ablation confirms that the effectiveness of MM-R2 comes not from any single isolated component, but from the combination of explicit intent grounding, structured retrieval-space selection, and iterative evidence acquisition.
% \vspace{-6pt}

\begin{figure*}[t]
    \centering
    \includegraphics[width=1\textwidth]{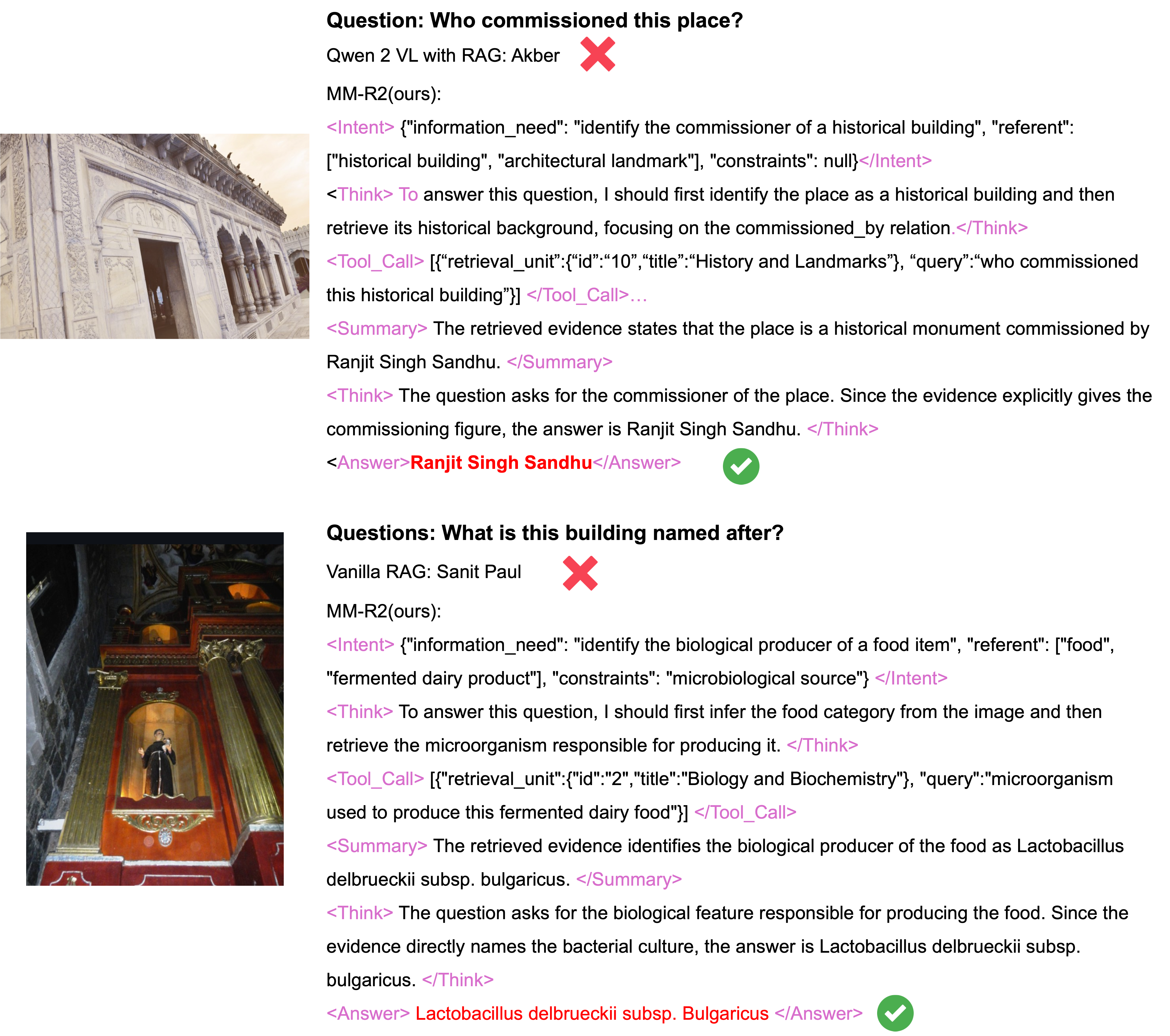}
    \caption{\textbf{Additional qualitative case comparisons.} In each example, the comparison method (or the preceding prediction) produces an incorrect answer, while MM-R2 predicts the correct answer through a structured retrieval trajectory with explicit intent grounding, retrieval actions, evidence summarization, and final answer generation. These cases illustrate how MM-R2 benefits from explicit retrieval planning and evidence integration.}
    \label{fig:case_study_appendix}
\end{figure*}

\subsection{Case Study}
% \vspace{-6pt}
% Additional qualitative case comparisons are provided in Appendix~\ref{app:case_study}.
% Additional qualitative case studies are provided in Appendix~\ref{app:case_study}, comparing MM-R2 against Qwen2-VL. The examples further show that MM-R2 supports more accurate retrieval routing, more coherent reasoning over retrieved evidence, and more evidence-grounded final predictions.
% % \vspace{-7pt}

To further illustrate the benefit of explicit retrieval planning, we present additional qualitative comparisons in Figure~\ref{fig:case_study_appendix}. In each case, the comparison method produces an incorrect answer, while MM-R2 arrives at the correct answer through a structured retrieval trajectory.

These examples highlight the key advantage of MM-R2: instead of predicting directly from loosely coupled multimodal signals, MM-R2 performs explicit intent grounding, structured routing, evidence retrieval, summarization, and answer generation. This process helps reduce ambiguity and spurious associations.

Overall, the examples suggest that MM-R2 benefits from three factors: explicit intent grounding, KnowledgeMap-based search-space selection, and evidence-based intermediate summarization. Together, these components enable more accurate and better-grounded answers.
% \begin{figure*}[t]
%     \centering
%     \includegraphics[width=1\textwidth]{fig/7.png}
%     \caption{\textbf{Additional qualitative case comparisons.} In each example, the comparison method (or the preceding prediction) produces an incorrect answer, while MM-R2 predicts the correct answer through a structured retrieval trajectory with explicit intent grounding, retrieval actions, evidence summarization, and final answer generation. These cases illustrate how MM-R2 benefits from explicit retrieval planning and evidence integration.}
%     \label{fig:case_study_appendix}
% \end{figure*}
\section{Conclusion}% \vspace{-6pt}
We introduced MM-R2, a multimodal agentic retrieval framework that explicitly models what to retrieve and where to search before retrieval. MM-R2 builds an intent-grounded retrieval state and performs structured, iterative retrieval over a KnowledgeMap. Trained with MM-R2-Traj, a large-scale trajectory dataset, MM-R2 improves answer accuracy on InfoSeek and Encyclopedic VQA while producing more interpretable and verifiable retrieval trajectories.

\bibliography{colm2026_conference}
\bibliographystyle{colm2026_conference}

\appendix
\section*{Appendix}

% \addcontentsline{toc}{section}{Appendix}

In this appendix, we provide case studies in
 Appendix~\ref{app:case_study}, additional details of the experimental setup in Appendix~\ref{setup}, the two-stage training pipeline in Appendix~\ref{app:two_stage_training}, and the trajectory synthesis process in Appendix~\ref{app:trajectory_synthesis}. We further present the prompt templates for the Intent-Referent Binding Module in Appendix~\ref{app:intent_binding_prompt}, additional details of KnowledgeMap construction and metadata generation in Appendix~\ref{app:knowledgemap_details}, the routing and query generation policy in Appendix~\ref{app:routing_prompt}, and the detailed definitions of our two-axis evaluation protocol in Appendix~\ref{evl}.

\section{Additional Details of Two-Stage Training}
\label{app:two_stage_training}

This appendix provides additional details of the two-stage training pipeline shown in Figure~\ref{fig3}. MM-R2 is first initialized by supervised fine-tuning (SFT) on MM-R2-Traj, and is then further optimized with Group Relative Policy Optimization (GRPO) to improve retrieval planning under task-level feedback.

\subsection{Stage 1: Supervised Fine-Tuning (SFT)}
\label{app:sft_details}

Starting from the base multimodal language model, we perform supervised fine-tuning on MM-R2-Traj as a cold-start stage. The goal of this stage is to teach the model to generate well-formed multi-step retrieval trajectories, including retrieval-unit selection, grounded query generation, evidence summarization, and answer generation.

Each training example is serialized as a structured interaction sequence with explicit XML-style tags, including \texttt{<intent>}, \texttt{<think>}, \texttt{<tool\_call>}, \texttt{<tool\_response>}, \texttt{<summary>}, and \texttt{<answer>}. Since the content inside \texttt{<tool\_response>} $\ldots$ \texttt{</tool\_response>} is produced by the external retrieval environment rather than by the model itself, these observation tokens are masked out during loss computation. Accordingly, the SFT objective is computed only over model-generated tokens.

Let $x$ denote the input context, including the image, the question, the KnowledgeMap metadata, and the preceding interaction history, and let $y=(y_1,\dots,y_{|y|})$ denote the serialized target trajectory. The SFT objective is defined as
\begin{equation}
\mathcal{L}_{\mathrm{SFT}}
=
-\sum_{t \in \mathcal{M}}
\log \pi_{\theta}(y_t \mid x, y_{<t}),
\end{equation}
where $\mathcal{M}$ denotes the set of token positions corresponding to model-generated segments, including \texttt{<intent>}, \texttt{<think>}, \texttt{<summary>}, \texttt{<tool\_call>}, and \texttt{<answer>}.

In this way, SFT teaches the model the basic structure of the intended agentic retrieval workflow while treating tool outputs as external observations. After this stage, the model can generate valid and structured multi-step retrieval trajectories and serves as the initialization for the subsequent reinforcement learning stage.

\subsection{Stage 2: Group Relative Policy Optimization (GRPO)}
\label{app:grpo_details}

Starting from the SFT-initialized policy, we further optimize MM-R2 with Group Relative Policy Optimization (GRPO)~\citep{shao2024deepseekmath}. For each query, we sample a group of $G$ rollouts by interacting with the retrieval environment. Each rollout $o_i$ consists of the model-generated reasoning and retrieval actions together with the corresponding tool responses returned by the environment.

As in the SFT stage, optimization is applied only to model-generated tokens. We adopt the following token-level clipped objective:
\begin{equation}
\begin{aligned}
\mathcal{J}(\theta)
=
\mathbb{E}_{\text{data},\,\text{rollouts}}
\Bigg[
\frac{1}{\sum_{i=1}^{G}|o_i|}
\sum_{i=1}^{G}\sum_{t=1}^{|o_i|}
\min \Big(
r_{i,t}(\theta)\hat{A}_{i},
\operatorname{clip}\big(
r_{i,t}(\theta),\,1-\varepsilon_{\mathrm{low}},\,1+\varepsilon_{\mathrm{high}}
\big)\hat{A}_{i}
\Big)
\Bigg]
\end{aligned}
\label{eq:grpo_full}
\end{equation}
where $|o_i|$ denotes the number of model-generated tokens in rollout $i$, and the importance ratio is
\begin{equation}
r_{i,t}(\theta)
=
\frac{\pi_{\theta}(o_{i,t}\mid c_{i,t})}
{\pi_{\theta_{\mathrm{old}}}(o_{i,t}\mid c_{i,t})}.
\end{equation}
Here, $c_{i,t}$ denotes the decoding context at token position $t$, including the prompt, the previously generated tokens, and the retrieved tool responses observed so far.

The group-relative advantage is computed by normalizing rollout rewards within the sampled group:
\begin{equation}
\hat{A}_{i}
=
\frac{R_i-\mathrm{mean}(\{R_j\}_{j=1}^{G})}
{\mathrm{std}(\{R_j\}_{j=1}^{G})}.
\label{eq:group_adv}
\end{equation}
In our formulation, $\hat{A}_{i}$ is a rollout-level advantage shared across all model-generated tokens in rollout $i$.

Our GRPO training uses a rollout-level reward that jointly captures routing quality and final answer correctness:
\begin{equation}
R_i
=
\lambda_{\mathrm{route}} R_i^{\mathrm{route}}
+
\lambda_{\mathrm{ans}} R_i^{\mathrm{ans}}.
\label{eq:reward_total_appendix}
\end{equation}
Here, $R_i^{\mathrm{route}}$ evaluates the quality of retrieval routing throughout rollout $i$, while $R_i^{\mathrm{ans}}$ evaluates whether the final predicted answer matches the reference answer.

At each retrieval step $t$, the agent selects a branch of the KnowledgeMap and issues a grounded query within that branch. We assign a step-level routing score $r_{i,t}^{\mathrm{route}}$ according to whether the selected branch at step $t$ is appropriate, and aggregate these scores across the trajectory:
\begin{equation}
R_i^{\mathrm{route}}
=
\frac{1}{T_i}\sum_{t=1}^{T_i} r_{i,t}^{\mathrm{route}},
\label{eq:route_reward_appendix}
\end{equation}
where $T_i$ is the number of retrieval steps in rollout $i$. The answer reward $R_i^{\mathrm{ans}}$ is computed by comparing the final predicted answer with the reference answer.

During rollout generation, the model sequentially produces \texttt{<think>}, \texttt{<tool\_call>}, \texttt{<summary>}, and \texttt{<answer>} segments while interacting with the retrieval environment. After each \texttt{<tool\_call>}, the environment returns evidence wrapped by \texttt{<tool\_response>}, which is appended to the current interaction history and used in subsequent steps. Each rollout terminates when the model emits a final \texttt{<answer>} or reaches a predefined maximum retrieval budget.

Overall, the two-stage design combines the complementary strengths of imitation learning and reinforcement learning: SFT establishes the basic multi-step retrieval workflow, while GRPO further refines routing and evidence-use decisions under task-level supervision.

\section{Trajectory Synthesis Details}
\label{app:trajectory_synthesis}

This appendix provides additional details on the construction of MM-R2-Traj, including candidate trace construction, teacher-based trajectory normalization, and quality filtering.

To synthesize retrieval trajectories at scale, we use the training splits of InfoSeek and Encyclopedic VQA as the source of multimodal question-answer examples. For each image-question pair $(I,Q)$, we first derive a structured intent state $z=(i,r,c)$ using the Intent-Referent Binding Module. Based on $(I,Q,z)$ and the available KnowledgeMap metadata, we then construct candidate retrieval traces that specify a retrieval unit, a grounded retrieval query, the retrieved evidence, and the resulting answer prediction.

A teacher multimodal reasoning model is subsequently used to convert these candidate traces into the structured textual format used by MM-R2. In particular, the teacher produces a standardized trajectory with explicit intent states, intermediate reasoning, tool calls, evidence summaries, and final answers. In this stage, the teacher primarily serves as a \emph{trajectory normalizer} that reformats and refines retrieval traces into a consistent supervision format, rather than as an answer-conditioned planner.

In our implementation, the teacher prompt is written in English and uses the image as input directly. The teacher is provided with the question, the image, the intent state, the available KnowledgeMap units, and the retrieved evidence associated with the candidate trace. These inputs preserve the multimodal grounding and retrieval context needed for trajectory construction, while keeping the synthesis process aligned with the retrieval setting faced by the agent at training time.

The raw teacher outputs are first collected in a structured reasoning format and then normalized into the same XML-style textual representation used by MM-R2 for training. Concretely, the final serialized trajectory uses explicit tags such as \texttt{<intent>}, \texttt{<think>}, \texttt{<tool\_call>}, \texttt{<tool\_response>}, \texttt{<summary>}, and \texttt{<answer>}. This conversion ensures consistency between synthesized trajectories and the downstream training format of the agent, while preserving the intermediate reasoning and evidence integration behavior that the model must learn.

To ensure data quality, we apply both automatic filtering and manual inspection. In the automatic stage, we remove trajectories with incorrect final answers, incomplete retrieval steps, inconsistent summaries, or degenerate reasoning patterns. Here, the reference final answer is used only for \emph{post-hoc verification and filtering}, rather than for authoring the free-form reasoning trace itself. We further manually inspect randomly sampled examples to verify retrieval-unit selection quality, query grounding, rationale faithfulness, and evidence support. After filtering, the final MM-R2-Traj dataset contains 900K trajectories, among which 22K involve two or three retrieval steps. The average serialized trajectory length is 2949 tokens, and the final split contains 860K training trajectories and 40K validation trajectories.
\begin{tcolorbox}[
    colback=gray!5,
    colframe=black!60,
colframe=orange!60!black,
    title={Prompt Template for Trajectory Synthesis}
]
\textbf{Role:} You are a multimodal retrieval-planning teacher.

\textbf{Task:} Given an image-question pair, an intent-grounded retrieval state, and the available KnowledgeMap units, generate a complete retrieval trajectory for solving the question.

\textbf{Known Inputs:}
\begin{itemize}
    \item \textbf{Question:} \texttt{\{question\}}
    \item \textbf{Image Path:} \texttt{\{image\_path\}}
    \item \textbf{Intent State:} \texttt{\{intent\_state\}}
    \item \textbf{Available KnowledgeMap Units:} \texttt{\{all\_knowledgemap\_units\}}
    \item \textbf{Selected KnowledgeMap Unit:} \texttt{\{selected\_unit\}}
    \item \textbf{Wikipedia ID:} \texttt{\{wiki\_id\}}
    \item \textbf{Wikipedia Title:} \texttt{\{wiki\_title\}}
    \item \textbf{Wikipedia Summary:} \texttt{\{wiki\_summary\}}
    \item \textbf{Retrieved Evidence:} \texttt{\{retrieved\_evidence\}}
\end{itemize}

\textbf{Instruction:}
Generate a structured retrieval trajectory that reflects how an agent should solve the question through multi-step retrieval. The trajectory should:
\begin{enumerate}
    \item reason about the information need and grounded visual referent;
    \item decide which KnowledgeMap unit(s) to access;
    \item issue grounded retrieval queries;
    \item inspect and summarize the returned evidence;
    \item determine whether additional retrieval is needed;
\end{enumerate}

\textbf{Output Format:}
Return the trajectory using the following XML-style tags only:
\texttt{<intent>}, \texttt{<think>}, \texttt{<tool\_call>}, \texttt{<tool\_response>}, \texttt{<summary>}, and \texttt{<answer>}.

Do not output any additional commentary outside the structured trajectory.
\end{tcolorbox}

Some example trajectories are shown below:

\label{app:routing_prompt}
\begin{tcolorbox}[
    breakable,
    colback=gray!5,
    colframe=black!60,
    title={Example Trajectory in MM-R2-Traj}
]
\small
\textbf{Question:} What is the immediate parent taxon of this plant? \\
\textbf{Image Path:} \texttt{oven\_04229931.jpg} \\
\textbf{Retrieval Units:} \texttt{[\{index, title, description\}, ...]}

% \vspace{4pt}
\texttt{<intent>} \\
\{"information\_need": "identify immediate parent taxon", "referent": ["plant", "parsley-like herb"], "constraints": null\} \\
\texttt{</intent>}

% \vspace{4pt}
\texttt{<think>} \\
To answer this question, I should first identify the plant species and then search for the taxonomic level directly above the species. \\
\texttt{</think>}

% \vspace{4pt}
\texttt{<tool\_call>} \\
\texttt{[{"retrieval\_unit":{"id":"4","title":"Biology"}, "query":"parsley scientific name and taxonomic hierarchy"}]} \\
\texttt{</tool\_call>}

% \vspace{4pt}
\texttt{<tool\_response>} \\
Parsley is \textit{Petroselinum crispum}, a species in the genus \textit{Petroselinum}. \\
\texttt{</tool\_response>}

% \vspace{4pt}
\texttt{<summary>} \\
The plant is parsley (\textit{Petroselinum crispum}). Its genus is \textit{Petroselinum}. \\
\texttt{</summary>}

% \vspace{4pt}
\texttt{<think>} \\
The evidence identifies the species, so the immediate parent taxon is its genus. Therefore, the answer is \textit{Petroselinum}. \\
\texttt{</think>}

% \vspace{4pt}
\texttt{<answer>} \\
Petroselinum \\
\texttt{</answer>}
\end{tcolorbox}

\label{app:routing_prompt}
\begin{tcolorbox}[
    breakable,
    colback=gray!5,
    colframe=black!60,
    title={Example Trajectory in MM-R2-Traj}
]
\small
\textbf{Question:} What is the architectural style of this place? \\
\textbf{Image Path:} \texttt{oven\_00018148} \\
\textbf{Retrieval Units:} \texttt{[\{id, caption\}, ...]}

% \vspace{4pt}
\texttt{<intent>} \\
\{"information\_need": "identify architectural style", "referent": ["Western Wall", "architecture", "Jerusalem"], "constraints": null\} \\
\texttt{</intent>}

% \vspace{4pt}
\texttt{<think>} \\
To answer this question, I should identify the historic site and retrieve evidence about its architectural background or construction period. \\
\texttt{</think>}

% \vspace{4pt}
\begin{verbatim}
<tool_call>
[{"retrieval_unit":{"id":"8","title":"Buildings"},
  "query":"architectural style of the Western Wall"}]
</tool_call>
\end{verbatim}

% \vspace{4pt}
\texttt{<tool\_response>} \\
The Western Wall is an ancient limestone wall in the Old City of Jerusalem, originally erected as part of the expansion of the Second Jewish Temple begun by Herod the Great. The lower courses are Herodian. \\
\texttt{</tool\_response>}

% \vspace{4pt}
\texttt{<summary>} \\
The Western Wall is an ancient limestone structure from the Second Temple period and is associated with Herodian construction. \\
\texttt{</summary>}

% \vspace{4pt}
\texttt{<think>} \\
The retrieved evidence links the Western Wall to Herodian construction during the Second Temple period. This corresponds to ancient Roman architecture. \\
\texttt{</think>}

% \vspace{4pt}
\texttt{<answer>} \\
ancient Roman architecture \\
\texttt{</answer>}
\end{tcolorbox}

\section{Intent-Referent Binding Module: Prompt Template}
\label{app:intent_binding_prompt}

To instantiate the Intent-Referent Binding Module $\Phi$, we use a fixed prompting-based multimodal language model to map each image-question pair $(I,Q)$ into a structured triplet $(i,r,c)$, where $i$ denotes the information need, $r$ denotes the grounded visual referent, and $c$ denotes task-specific retrieval constraints. In practice, the prompt is designed to explicitly identify what knowledge is being asked for, which image-grounded entity or region the query refers to, and what additional constraints should guide subsequent retrieval. During large-scale preprocessing, we useimages as lightweight substitutes for raw image inputs, which makes annotation more efficient while preserving the core visual semantics needed for intent grounding.

\begin{tcolorbox}[
    breakable,
colback=purple!5,
colframe=purple!50!black,
    boxrule=0.6pt,
    before skip=6pt,
    after skip=6pt,
    title={Prompt Template for Intent-Referent Binding}
]
\textbf{Role:} You are an intent-grounding module for multimodal retrieval.

\textbf{Task:} Given a question and animage, infer a structured intent state that captures the information need, the grounded visual referent, and any task-specific retrieval constraints.

\textbf{Known Inputs:}
\begin{itemize}
    \item \textbf{Question:} \texttt{\{question\}}
    \item \textbf{Image Path:} \texttt{\{image\_path\}}
\end{itemize}

\textbf{Instruction:}
Infer a structured intent representation for the multimodal query.
The output should identify:
\begin{enumerate}
    \item the \textbf{information need}, i.e., what knowledge the question is asking for;
    \item the \textbf{referent}, i.e., the image-grounded object, entity, scene element, or region that the question concerns;
    \item the \textbf{constraints}, i.e., answer type, granularity, temporal scope, geographic scope, or any cross-modal inconsistency that should constrain retrieval.
\end{enumerate}

\textbf{Output Format:}
Return \textbf{only} a single valid JSON object with the following structure:
\begin{verbatim}
{
  "intent": {
    "information_need": "<concise description of the knowledge being asked for>",
    "referent": ["<grounded entity or concept>", "..."],
    "constraints": null OR "<short description of retrieval constraints>"
  }
}
\end{verbatim}

\textbf{Requirements:}
\begin{itemize}
    \item Use concise phrases rather than long explanations.
    \item If there are no special constraints, set \texttt{"constraints"} to \texttt{null}.
    \item If there is a mismatch between the question and the visual content, express it briefly in \texttt{"constraints"}.
    \item Do not output markdown, comments, or additional text outside the JSON object.
\end{itemize}
\end{tcolorbox}

The resulting structured intent state is then normalized into the triplet $(i,r,c)$ used by MM-R2, and serves as a fixed conditioning signal for subsequent routing and within-unit query generation.

\section{Additional Details of KnowledgeMap Construction}
\label{app:knowledgemap_details}

We provide additional details on the offline construction of the KnowledgeMap used in MM-R2. In our framework, the KnowledgeMap serves as a structured routing scaffold over the external corpus: it makes the \emph{where-to-search} decision explicit, while the final evidence retrieval is still performed at the passage level within the selected retrieval unit(s).

\paragraph{KnowledgeMap construction.}
Starting from the InfoSeek knowledge base, we first encode each Wikipedia-derived passage into a dense semantic embedding space. We then perform automated search over HDBSCAN configurations to identify a clustering setting that yields semantically coherent and reasonably balanced retrieval units. The final configuration is selected by jointly considering semantic coherence, cluster balance, assignment coverage, and downstream retrieval utility. Under the selected configuration, the corpus is partitioned into 10 retrieval units, which define the KnowledgeMap used in MM-R2.

After clustering, each passage is assigned to one primary retrieval unit, yielding an approximate partition of the corpus. To make these units interpretable for routing, we further generate lightweight metadata for each unit, including a short title and a brief natural-language description summarizing its dominant content. We also apply quality-checking procedures to refine noisy, imbalanced, or semantically incoherent units.

At a high level, the resulting 10 retrieval units cover major semantic areas of the corpus, such as sports and competition, history and conflict, politics and society, geography and places, transport and technology, biology and life sciences, literature and the arts, music, screen media, and academic knowledge. For example, the sports-related unit contains entries on Olympic events, professional tournaments, athlete biographies, and competition results; the biology-related unit groups species, taxonomy, biodiversity, and life-science concepts; and the geography-related unit covers settlements, natural features, and landmarks. This organization is not intended to define a perfect ontology over the corpus, but rather to provide a practical and interpretable routing space for downstream retrieval.

\paragraph{Metadata generation.}
To support routing, each retrieval unit is associated with lightweight metadata generated by a large language model. Given representative passages sampled from the same cluster, the model produces a concise description that captures the dominant topic of that unit. These metadata are used only as a routing interface for the agent and are not treated as evidence themselves. The actual evidence remains the underlying passages assigned to the selected retrieval unit.

An example metadata entry is shown below:

\begin{tcolorbox}[breakable, colback=blue!5,colframe=blue!50!black, title=Example Metadata of a Retrieval Unit]
\small
\begin{verbatim}
{
  "id": "9",
  "title": "Music"
  "description": "Artists, albums, and songs across rock, country, jazz, pop, and more."
}
\end{verbatim}
\end{tcolorbox}

This example corresponds to a music-oriented retrieval unit. Similar metadata are generated for all units to provide the agent with an interpretable summary of the available routing space before retrieval.

\paragraph{Embedding and retrieval indexing.}
For each retrieval unit, we build an independent dense retrieval index. Concretely, we concatenate the Wikipedia title, summary, and truncated content for each entry, split the text into overlapping chunks, and encode them using \texttt{BAAI/bge-large-en} implemented with \texttt{SentenceTransformer}. We use normalized dense embeddings, so retrieval is performed with inner-product similarity in FAISS. In our implementation, the chunk size is 800 characters with 100-character overlap, the content is truncated to 1600 characters, and the encoder batch size is set to 1024. We build a separate \texttt{IndexFlatIP} index for each retrieval unit, preferably on GPU and then serialize it as a CPU index for later retrieval.

\section{Routing and Query Generation Policy}
\label{app:routing_prompt}

As described in Section~\ref{subsec:policy}, at each retrieval step the agent first selects a retrieval unit from the KnowledgeMap and then formulates a grounded retrieval query within that unit. We implement this behavior using a dedicated prompt that isolates the \emph{where-to-search} and \emph{what-to-search} decisions before evidence is returned by the retriever.

\paragraph{Prompt design.}
Given the multimodal input $(I,Q)$, the structured intent state $z=(i,r,c)$ produced by the Intent-Referent Binding Module, and the list of candidate retrieval units in the KnowledgeMap, the agent is prompted to generate two outputs: (1) a natural-language reasoning segment that explains which retrieval unit is most relevant to the current question and image, and (2) a structured tool call specifying the selected retrieval unit id together with the generated retrieval query. Importantly, at this stage the agent does not observe any retrieved evidence and therefore does not produce summaries or final answers. This design ensures that routing decisions are made purely from the multimodal query, the intent state, and the semantic descriptions of the available retrieval units.

\paragraph{Inputs and outputs.}
At step $t$, the prompt includes the image $I$, the question $Q$, the intent state $z=(i,r,c)$, and a compact representation of the candidate retrieval units, where each unit is represented by its identifier and metadata description. The agent is required to output:
\begin{itemize}
    \item \texttt{<tool\_call>}: a JSON-formatted retrieval action of the form
    \begin{verbatim}
[{"retrieval_unit":{"id":"..."},"query":"..."}]
    \end{verbatim}
    where \texttt{id} must exactly match one valid retrieval-unit id from the provided candidate list, and \texttt{query} must be a concise retrieval-oriented query grounded in $(I,Q,z)$.
\end{itemize}

This prompt structure operationalizes the policy described in Section~\ref{subsec:policy}: the routing action $a_t$ corresponds to retrieval-unit selection, while the generated query corresponds to $q_t$. The pair $(a_t, q_t)$ is then executed by the external retrieval backend to obtain evidence $e_t$.

\paragraph{Prompt template.}
A simplified version of the routing and query generation prompt is shown below.

\begin{tcolorbox}[
    breakable,
colback=green!5,
colframe=green!40!black,
    title={Prompt Template for Retrieval Unit Selection and Query Generation}
]
\textbf{Role:} You are a multimodal retrieval agent.

\textbf{Goal:}  
Given the user’s question, image, an intent-grounded retrieval state, and a list of candidate retrieval units, your task in this step is only to:
\begin{enumerate}
    \item select the most relevant retrieval unit id;
    \item construct a retrieval query for that unit.
\end{enumerate}

\textbf{Inputs:}
\begin{itemize}
    \item \textbf{Question:} \texttt{\{question\}}
    \item \textbf{Image Path:} \texttt{\{image\_path\}}
    \item \textbf{Intent:} \texttt{\{intent\}}
    \item \textbf{All Retrieval Units:} \texttt{\{all\_retrieval\_units\}}
\end{itemize}

\textbf{Required Outputs:}
\begin{enumerate}
    \item \textbf{\texttt{<think>}}  
    Reason which single retrieval unit id is most relevant by aligning the question, image, and intent state with the semantic descriptions of the candidate retrieval units.

    \item \textbf{\texttt{<tool\_call>}}  
    A JSON array with one object:
\begin{verbatim}
[{"retrieval_unit":{"id":"..."},"query":"..."}]
\end{verbatim}
    where \texttt{"id"} must be one valid id from the provided candidate list, and \texttt{"query"} must be a concise retrieval-oriented query grounded in the multimodal input and the intent.
\end{enumerate}

\textbf{Rules:}
\begin{itemize}
    \item The question and image are the primary basis for the decision.
    \item The intent is used to clarify the information need, grounded referent, and retrieval constraints.
    \item The selected retrieval unit id must exactly match one id from the provided candidate list.
    \item Do not output \texttt{<summary>}, \texttt{<think\_answer>}, or \texttt{<answer>} in this step.
    \item Do not output extra text outside the required tags.
\end{itemize}
\end{tcolorbox}

\paragraph{Interaction with the retriever.}
After the agent emits \texttt{<tool\_call>}, the external retrieval tool executes the generated query only within the selected retrieval unit and returns the retrieved evidence as \texttt{<tool\_response>}. This retrieved evidence corresponds to $e_t$ in the main formulation. The tuple $(a_t,q_t,e_t)$ is then appended to the retrieval history $h_t$ and used as part of the interaction context for subsequent steps. In this way, multi-step retrieval is realized as an explicit loop of retrieval-unit selection, grounded query generation, evidence observation, and history-aware replanning.

\section{Two-Axis Evaluation for Agentic RAG}
\label{evl}
% \vspace{-1mm}
Because MM-R2 explicitly exposes intermediate retrieval decisions, we evaluate it along two complementary axes: answer-level effectiveness and process-level verifiability. This design is especially important for agentic multimodal RAG: beyond producing a correct final answer, the system should also make explicit where it searches, what evidence it relies on, and whether the exposed trace can genuinely justify the answer.
% \vspace{-1mm}
\noindent
\textbf{Axis 1: Task-level Effectiveness.}
This axis evaluates whether explicit pre-retrieval reasoning improves both retrieval and final answer quality. For answer quality, we report standard QA metrics on final answers, including Accuracy and F1 when applicable. For retrieval quality, we report Recall@k, which measures whether the gold evidence or target retrieval unit appears among the top-k retrieved results.
% \vspace{-1mm}

\noindent
\textbf{Axis 2: Process-level Verifiability.}
This axis asks whether the exposed retrieval trajectory reflects the intended behavior of MM-R2 as an agentic mRAG system: routing to the appropriate region of the KnowledgeMap, revealing concise supporting evidence, and making the final prediction reproducible from the exposed trace rather than hidden internal reasoning alone.
% \vspace{-3mm}
\begin{itemize}[leftmargin=*]
\item \textbf{LLM-as-a-Judge (LJ).}
This metric evaluates whether the exposed trajectory is a \emph{trustworthy explanation} of the model's behavior. A strong LLM inspects the user input, the agent's exposed trace, and the retrieved evidence, and assigns scores for (i) trace--answer consistency, namely whether the trajectory forms a coherent decision path toward the final answer, and (ii) evidence support, namely whether the cited evidence actually supports the answer without excessive noise or irrelevant content.

\item \textbf{Route Recall@1.}
This metric evaluates whether the agent learns \emph{where to search}, which is a central goal of MM-R2. Since MM-R2 replaces flat retrieval with explicit routing over the KnowledgeMap, a good agent should first identify the correct retrieval region before issuing grounded within-unit queries. Let $U_q^\star$ be the gold KnowledgeMap unit(s) for query $q$, and let $\hat{U}_q$ be the route predicted by the agent. We define
\begin{equation}
\text{RouteRecall@1}
=
\frac{1}{|Q|}
\sum_{q \in Q}
\mathbf{1}\!\left[\hat{U}_q \cap U_q^\star \neq \emptyset\right].
\end{equation}

\item \textbf{Explanation Sparsity.}
This metric evaluates whether the agent exposes \emph{selective and focused} evidence rather than verbose or noisy summaries. In agentic retrieval, long explanations are not necessarily better; ideally, the model should preserve only the evidence that is most relevant to the final decision. Let $E_q$ denote the evidence summary for query $q$, and let $|E_q|$ denote its token length. We define
\begin{equation}
\text{Explanation Sparsity}
=
\frac{1}{|Q|}
\sum_{q \in Q} |E_q|,
\end{equation}
where lower values indicate more concise evidence exposure. We interpret this metric jointly with support and fidelity, so that shorter summaries are preferred only when they still preserve the information needed for justified answering.

\item \textbf{Fidelity.}
This metric evaluates whether the final prediction is \emph{actually grounded in the exposed evidence}. In other words, it asks whether the answer can be reproduced from the revealed evidence summary, instead of depending on hidden internal knowledge or unrevealed reasoning steps. We compare the accuracy of the full agent pipeline, $\text{Acc}_{\text{full}}$, with that of a variant that answers each question using only the exposed evidence summary (with the same backbone), $\text{Acc}_{\text{evid}}$. We define
\begin{equation}
\text{Fidelity}
=
\frac{\text{Acc}_{\text{evid}}}{\text{Acc}_{\text{full}}}.
\end{equation}
If fidelity remains high, the model's behavior is largely reproducible from the revealed evidence; if it drops sharply, the model is relying on information outside what the trace claims as justification.
\end{itemize}

\end{document}